\documentclass[screen]{acmart}

\usepackage{multirow}
\usepackage{longtable}
\usepackage{rotating}
\usepackage{afterpage}
\usepackage{color,soul}

\AtBeginDocument{%
  \providecommand\BibTeX{{%
    \normalfont B\kern-0.5em{\scshape i\kern-0.25em b}\kern-0.8em\TeX}}}

\setcopyright{acmcopyright}
\copyrightyear{2021}
\acmYear{2021}
\acmDOI{XXX}

\begin{document}

%%
%% The "title" command has an optional parameter,
%% allowing the author to define a "short title" to be used in page headers.
\title{Prose2Poem: The Blessing of Transformers in Translating Prose to Persian Poetry}

%%
%% The "author" command and its associated commands are used to define
%% the authors and their affiliations.
%% Of note is the shared affiliation of the first two authors, and the
%% "authornote" and "authornotemark" commands
%% used to denote shared contribution to the research.
\author{Reza Khanmohammadi}
\authornote{Both authors contributed equally to the paper}
\email{rezanecessary@gmail.com}
\orcid{0000-0002-6484-6983}
\affiliation{%
  \institution{University of Guilan}
  \country{Iran}
}

\author{Mitra Sadat Mirshafiee}
\authornotemark[1]
\email{mitra.mirshafiee@gmail.com}
\orcid{https://orcid.org/0000-0002-7108-645X}
\affiliation{%
  \institution{Alzahra University}
  \country{Iran}
}

\author{Yazdan Rezaee Jouryabi}
\email{yazdanrezaeejouryabi@gmail.com}
\orcid{0000-0001-7500-1057}
\affiliation{%
  \institution{Shahid Beheshti University}
  \country{Iran}
}

\author{Seyed Abolghasem Mirroshandel}
\authornote{Corresponding Author}
\email{mirroshandel@guilan.ac.ir}
\orcid{0000-0001-8853-9112}
\affiliation{%
  \institution{University of Guilan}
  \country{Iran}
}

%%
%% By default, the full list of authors will be used in the page
%% headers. Often, this list is too long, and will overlap
%% other information printed in the page headers. This command allows
%% the author to define a more concise list
%% of authors' names for this purpose.

% \renewcommand{\shortauthors}{Trovato and Tobin, et al.}

%%
%% The abstract is a short summary of the work to be presented in the
%% article.
\begin{abstract}
Persian Poetry has consistently expressed its philosophy, wisdom, speech, and rationale based on its couplets, making it an enigmatic language on its own to both native and non-native speakers. Nevertheless, the noticeable gap between Persian prose and poem has left the two pieces of literature medium-less.  Having curated a parallel corpus of prose and their equivalent poems, we introduce a novel Neural Machine Translation (NMT) approach to translating prose to ancient Persian poetry using transformer-based language models in an exceptionally low-resource setting. Translating input prose into ancient Persian poetry presents two primary challenges: In addition to being reasonable in conveying the same context as the input prose, the translation must also satisfy poetic standards. Hence, we designed our method consisting of three stages. First, we trained a transformer model from scratch to obtain an initial translations of the input prose. Next, we designed a set of heuristics to leverage contextually-rich initial translations and produced a poetic masked template. In the last stage, we pretrained different variations of BERT on a poetry corpus to use the masked language modelling technique to obtain final translations. During the evaluation process, we considered both automatic and human assessment. The final results demonstrate the eligibility and creativity of our novel heuristically aided approach among Literature professionals and non-professionals in generating novel Persian poems.
\end{abstract}

%%
%% The code below is generated by the tool at http://dl.acm.org/ccs.cfm.
%% Please copy and paste the code instead of the example below.
%%


%%
%% Keywords. The author(s) should pick words that accurately describe
%% the work being presented. Separate the keywords with commas.

%%
%% This command processes the author and affiliation and title
%% information and builds the first part of the formatted document.
\maketitle

\section{Introduction}
The dawn of Persian Poetry dates back to almost 12 centuries ago when Rudaki, the founder of classical Persian/Farsi language and poetry, inaugurated a new literary pathway. Persian has been one of the most poetic languages throughout history, having been home to great poets, each contributing in their own style to its literature. According to Charles Baudelaire, a French poet, poetry serves as a bridge between the language of the universe and the universe of language. Historical events, emotions, reflections, objections, hopes, and dreams have been among the different aspects involved in the cultural identities of Persian poems. Hence, they are expected to convey rich and heavy loads of information by living up to Linguistic Aesthetics' expectations making poetry generation extremely challenging.

Persian poems cannot be easily comprehended and are hard to interpret and understand. Aside from its heavy literature, Figures of Speech such as \textit{Metaphor}, \textit{Simile}, \textit{Paradox}, and \textit{Hyperbole} can potentially exert pressure on what can be a bewildering literature even for native speakers and professionals \cite{articleMoradi, inbookw}. Moreover, the texture and presentation of poems is subjective, and the factors influencing its genesis are nonidentical among poets due to their individually unique rhetorical styles. In contrast to Persian prose, wherein grammar is defined structurally and is straightforward, poetry is accustomed to describing a conceptual idea using a creative language wherein sanely-employed irregular and anomalous speech is acceptable. In the world of Persian poetry, the more literary rules a poem wisely breaks, the more impressive it is.

The aforementioned challenges demonstrate the difficulty of understanding and generating Persian poetry among both professionals and non-professionals and evidently, machines are no exception. Poetry generation is not a straightforward task and requires a considerable amount of data for intelligent systems to learn how to write an acceptable poem. However, doing so is even more problematic in Persian since ancient poems are scarce and much more intricate than English poems \cite{elaheh2011poetry}. While a poet has to account for many poeticness constraints in the first, one experiences a relatively higher degree of freedom in generating the latter. As expected, fewer studies have addressed the Poetry Generation task in Persian, whereas various Recurrent and Transformer-based approaches and baselines have been introduced in English \cite{ghazvininejad-etal-2017-hafez, Xie2017DeepP, lau-etal-2018-deep}. Fortunately, pretrained Language Models have recently stepped forward in Persian Natural Language Processing \cite{farahani2020parsbert} and have paved the way for advanced methodologies to address the generative aspect of Persian Poetry \cite{ParsGPT2}.

In this work, we take the first step in translating prosaic Persian text to ancient Persian poetry by introducing a novel machine translation approach called \textbf{Prose2Poem}. Conventional deep neural machine translation techniques utilize a seq2seq architecture, namely encoder-decoder networks, to encode input text as latent features from which the decoder can extract the output text. Such networks require a parallel corpus to learn from, however, no parallel prose-poem dataset was available at the time of conducting this research. Hence, as the first step, we collected such a dataset and developed an encoder-decoder model. Nonetheless, the complex nature of Persian poetry makes it highly difficult for such networks to converge and generalize. More specifically, with the amount of data available, such networks fail to translate prose to a standard poem that conveys the same context as input. Despite our dataset containing only 5,191 parallel prose-poem samples, a relatively broader span of raw ancient poems is available (321,581 couplets). This provided us with the opportunity to develop a translation system that learns how to translate the context and generate standard poems through utilizing the parallel and the monolingual datasets, respectively. We used an attention-based seq2seq architecture for the first objective and the Masked Language Modelling (MLM) technique for the latter. To integrate the two, we came up with a set of heuristics trying to mimic the unique process of Persian poems generation. Furthermore, we designed our heuristics to be adaptable to generate translated poems in six different formats of ancient Persian poetry. The difference between these formats lies in the location of rhymes in the hemistichs of a couplet. Namely, these formats are \textit{Robaei}, \textit{Ghazal}, \textit{Ghasideh}, \textit{Masnavi}, \textit{Ghet'e}, and \textit{Dobeiti}.

Our contributions in this work are as follows:
\begin{enumerate}
    \item We established a firm baseline for the problem of Persian Prose to Poem translation by introducing a new method of low-resource NMT capable of better handling the unique nature of Persian poetry is developed.
    \item We introduced a parallel prose-poem corpus and a synonym-antonym dataset in Persian.
    \item We maked curated datasets and all implementation publicly available\footnote{https://github.com/mitramir55/Prose2Poem} to aid further studies.
\end{enumerate}

The rest of this paper is organized as follows: First, in section 2, we discuss the previous literature of Persian poetry and the bottlenecks of processing this natural language. Next, we break down our proposed NMT method in section 3. In section 4, we evaluate the results of our experimentation and provide several examples illustrating how our method performs on real-world circumstances. Finally, we conclude the paper in section 5.

\section{Previous Work}
The applicability of Natural Language Processing (NLP) techniques has long been investigated in Poetry. Still, Poetry is a difficult phenomenon to computationally process because of its complexity. Poems are extremely rich in context \cite{kesarwani-etal-2017-metaphor} and very confusing for the machine to represent them \cite{unknown19}. Predictably, a vast majority of initial studies had to first analyze poetry's core values such as Rhyme \cite{jhamtani-etal-2019-learning, hopkins-kiela-2017-automatically}, Meter \cite{Tanasescu2016AutomaticCO}, Word Similarities \cite{articleZ} and Stresses \cite{agirrezabal-etal-2016-machine} to gradually push the field towards better Poetic Machine Intelligence.

Poetry's generative aspect has also inspired a considerable amount of literature where various approaches have been explored over time \cite{agirrezabal-etal-2013-pos, Xie2017DeepP, lau-etal-2018-deep, Chen2019SentimentControllableCP, bena2020introducing, ghazvininejad-etal-2017-hafez, yi-etal-2018-automatic}, including statistical inference, LSTMs, and Variational Autoencoders. Additionally, as many other NLP research lines have witnessed, the revolution of Attention-based Language Models and Transformers has also paved the way for further advancements in this field. For instance, \citet{p1p} leveraged transformers to first generate a list of candidate poems. The list is filtered afterwards by calculating the cosine similarity between candidate poems and a Doc2Vec \cite{doc2vec} model to choose the most cohesive candidate as output. Also, using transformers, \citet{p2p} demonstrated the advantages of cross-lingual transfer between Spanish, German, and English in predicting the metrical pattern of poems. Other studies have addressed Poetry Generation in other interesting settings. \citet{Liu_2018} and \citet{xu2018images} introduced approaches to generating poetry based on visual clues in English and Chinese, whereas \cite{LollerAndersen2018DeepLP} developed the same concept while seeking to better satisfy rhythmical expectations.

Generally, translating prose to poem has attracted minor attention, and very few previous studies have addressed this problem. \citet{gokirmak2021converting} has recently studied the applicability of back-translation \cite{backtranslation}, transformer-based language models, and GPT-2 \cite{Radford2019LanguageMA} in translating prose to poem. The study utilized a Czech, a Turkish, and an English corpus with around 183k, 670k, and 191k poems, respectivelly. These numbers are significantly larger than the single available dataset we used in our study, which sentences our work to be conducted under extremely low-resource setting. Fortunately, a plethora of prior studies have investigated means of integrating monolingual data into machine translation tasks. For instance, Cold Fusion \cite{sriram2017cold}, Shallow Fusion \cite{gulcehre2015using}, and Deep Fusion \cite{gulcehre2015using} have engaged monolingual language models with NMT architectures to enhance translation quality. These articles inspired us to initiate a similar approach on Persian language.

\begin{figure*}[!t]
\begin{center}
\includegraphics[width=\linewidth]{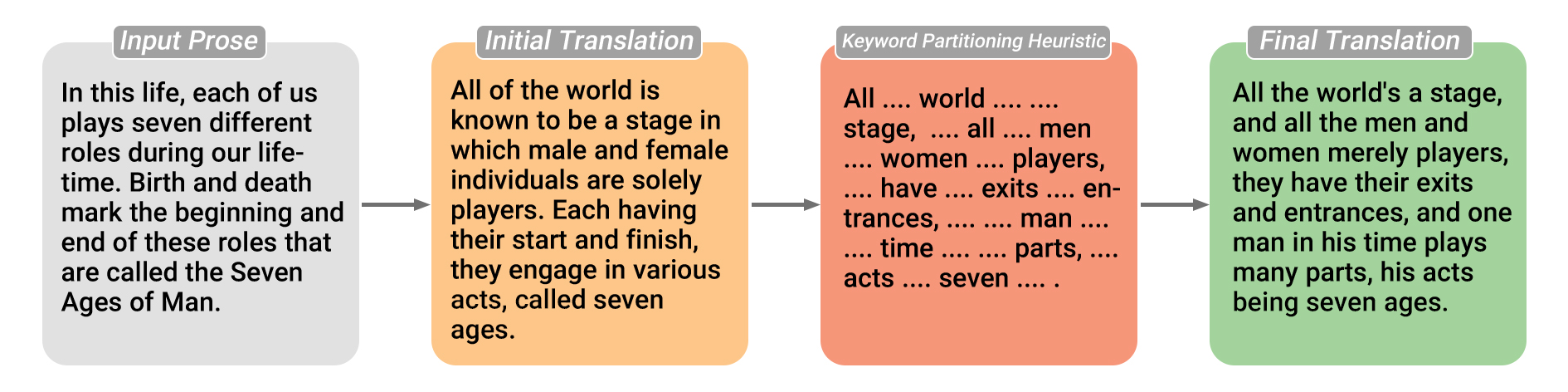}
\end{center}
\caption{A schema of how input is processed in each stage (Note: an English example is provided to aid non-native readers. Whereas our proposed method solely focuses on Persian)}
\label{fig:gsteps}
\end{figure*}

\begin{figure*}[!t]
\begin{center}
\includegraphics[width=0.9\linewidth]{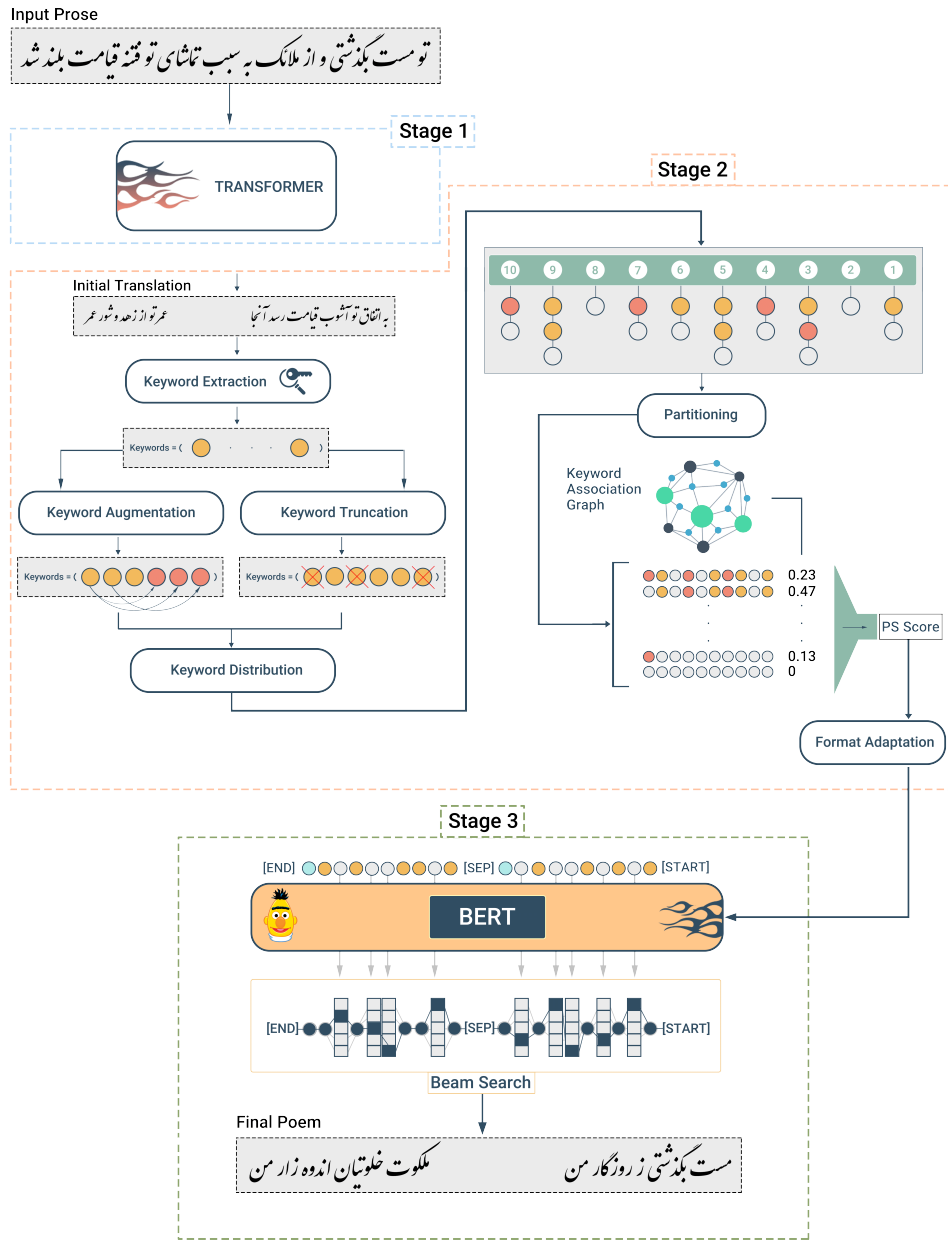}
\end{center}
\caption{An overview of our final approach}
\label{fig:stages}
\end{figure*}

\section{Proposed Method}
The performance of machine translation models is heavily reliant on parallel datasets, and with no such Persian datasets available, we found this an excellent opportunity to gather a parallel set of prose and poems to tackle our problem.  Although valid poetry sources are exceptionally rare, tthe Didarejan\footnote{https://didarejan.com/} website has matched simple Persian prose to semantically similar Ghazal poems written by ancient Persian poets, such as \textit{Hafez}, \textit{Saadi}, and \textit{Molana} (\textit{Rumi}). Our first step towards developing a machine translation method involves extracting these parallel prose and poems as our dataset of interest.
% the Ganjoor\footnote{www.ganjoor.net} dataset, which is a source of Persian poems in chronological order

We break down our translation task into three stages. First, in stage 1, we generate an Initial translation of the input prose using conventional NMT approaches. Due to data scarcity, outputs at this stage contain poetic flaws and shortcomings. To complement our initial results, we process initially-translated outputs for two additional stages. To accomplish this, we designed a heuristic in stage 2 to map initial translations to a set of masked sequences from which pretrained MLMs will then generate Final Translations in stage three (Figure \ref{fig:gsteps}). Since machine translation tasks deal with text generation, generative language models (e.g. GPT2) seem to be a perfect component to be incorporated. Whereas, this study outlines the applicability of MLMs (e.g. BERT \cite{BERT}) in generating poetry. Combining this technique with a beam search algorithm lets us control a generative setting in which limitations and customizations can be applied to produce outputs that better suit poetic needs.

\subsection{Stage 1: Initial Translation}
In terms of standard architectures used among translation models, encoder-decoders are among the oldest contributing architectures to sequence-to-sequence (Seq2Seq) problems. Besides, when incorporated with the attention mechanism, such models can also overcome the challenge of decoding long input sequences and vanishing gradients. As the most pivotal component of the revolutionizing Transformer architecture, applying this mechanism multiple times, one allows for a wider range of attention span. Having mentioned the advantages of leveraging this mechanism, we initially tackle the translation task by training a Recurrent model, an Attention-based Encoder-Decoder model, and a Transformer-based Encoder-Decoder model on our parallel corpus. Given set \(X_{1:a} = \{x_1, ..., x_a\}\) as an input sequence (prose) with \(a\) length of words, we have defined stage 1 Seq2Seq models as functions that translate \(X_{1:a}\) input words to \(I_{1:m}\) target words (Equation \ref{eq:1}).

\begin{equation}
\label{eq:1}
    Stage \: 1: X_{1:a} \rightarrow I_{1:m}
\end{equation}
The most striking observation about encoder-decoder models is their effectiveness in expressing prose context as poetry. In other words, encoder-decoders' generated poems bear a close semantic resemblance to both the input prose and the ground-truth poem. In addition, ancient poems are generally context-rich, and so the ability to include context within generated poems is extremely valuable. Additionally, these models learn to generate poems with fair lengths and acceptable poetic meters. The outputs, however, often lack poeticness, which means that they do not adhere to particular literary rules and cannot be considered standard Persian poetry. Below is a list of their shortcomings:
\begin{itemize}
  \item Very few generated samples in set \(I_{1:m}\) seem to have formed rhythmic literature.
  \item Despite exhibiting an appropriate level of context transformation from \(X_{1:a}\) to \(I_{1:m}\), Seq2Seq models didn't seem to \textit{narrate} that information and arrive at a precise meaningful point in our experiments.
  \item Set \(I_{1:m}\) forms irrational associations of words. 
\end{itemize}
Consequently, encoder-decoder architectures produce incoherent and unintelligible results with weak human evaluation scores on our dataset, making them an inefficient alternative. Training machine translation models with small corpora has the disadvantage that even robust architectures fail to generalize well and overfit the data. As our setting is extremely low-resourced, the model cannot simultaneously learn from insufficient data to map input features to output features and also produce well-written ancient poems, which alone is an arduous task. Alongside our trained NMT models, we also experimented with two available pretrained Persian GPT2 models from \citet{bolbol} and \citet{ParsGPT2} that were already trained on the Ganjoor dataset. Still, these approaches either failed to converge or were stuck in local minima or produced low-quality poems. Having mentioned the encoder-decoder models' inefficiencies in poem generation, the scarcity of our parallel data, the uniqueness of Persian poetry's nature, and the availability of a relatively larger number of monolingual data on the target side, we introduced the following two additional stages.

\subsection{Stage 2: Keyword Partitioning Heuristic}
As shown in Figure 1, the Keyword Partitioning Heuristic receives Initial Translation's contextually-rich outputs and generates masked sequences. In several ancient poems, poets have written more than one couplet to express the ideas they had in mind. Hence, we design the heuristic adaptable to generating multiple masked sequences. Given the initial translation of an input prose, we define heuristic \(h\) (Equation \ref{eq:2}) as a function that generates \(c\) masked sequences \(M_{1:c}\).
% (\verb|[MASK]|)
\begin{equation}
\label{eq:2}
    h:I \rightarrow M_{1:c}
\end{equation}

Generated masked sequences contain a set of tokens that are either keywords or masks. These keywords represent the context of initial translations which, as previously mentioned, are well-captured by encoder-decoders in stage 1. Besides, masked tokens are placeholders that will be further predicted using MLMs in stage 3. This is while stage 2, is all about forming a poetic template that well captures initially-translated poems' context and provides a proper infrastructure for MLM to generate final poems. Evidently, the quality of our final outcome is heavily dependent on the quality of our keyword/mask partitioning. Below, we define a list of criteria we expect \(h\) to consider:
\begin{itemize}
    \item \textbf{Keywords Presence:} A fair number of keywords should appear in a masked sequence. This way, keywords better guide stage 3 MLM to generate a high-quality poem through mask prediction.
        \item \textbf{Keywords Adjacency:} To encourage the generative aspect of our work, keywords should be rationally separated, and enough mask spacing should be present among adjacent keywords.
    \item \textbf{Keywords Association:} Persian Poetry's Literature is fulfilled with highly literary collocations. Hence, to understand and capture these collocations, \(h\) should avoid partitions in which keywords are nonliterary associated.
\end{itemize}
Henceforth, we step through different components of \(h\) to generate \(c\) masked sequences while striving to meet the previously mentioned criteria. An illustration of the following components is provided in Figure \ref{fig:stages}.

\subsubsection{Keyword Extraction}
Primarily, we extract keywords from the contextually rich \(I\) set. We used an unsupervised keyword extraction package called YAKE! \cite{CAMPOS2020257} to automatically detect keywords in our text. This unbiased method is neither task nor language-specific, making it suitable for our work. stage 1 encoder-decoder models have successfully learned how to decode couplet-like outputs that is composed of two hemistichs. We feed the I set to YAKE! to capture a list of couplet-level keywords based on the big picture that the poem draws. Subsequently, capturing as much context as possible, we feed each hemistich individually to the model and extract hemistichs-level keywords. Finally, we combine the two lists and define the final set of keywords as \(K_{1:n}\). Since some keywords can be extracted more than once among the two levels, we define set \(F^{k}_{1:n}\) to keep track of each keyword's frequency of appearance.

\subsubsection{Keyword Augmentation/Truncation}
To overcome the challenge of Keywords Presence, we augment the \(K\) set to highlight context information. We gathered a Synonyms-Antonyms Dataset for all 4,756 words in the Ganjoor Dataset. Given function \(Syn(K_{i}, j)\), which returns the \(j\)\textsuperscript{th} most frequent synonym of \(K_{i}\), we define the synonyms set (\(S\)) and the Final Keywords set (FK) as below.
\begin{equation}
\label{eq:9}
    S = \{ Syn(K_i, j) | 1 \leq i \leq n , 1 \leq j \leq F^k_i \}
\end{equation}
\begin{equation}
\label{eq:10}
    FK_{1:n^{\prime}} = K \cup S
\end{equation}

Having defined the \(FK\) set, we now define \(c\), which was previously used to declare the number of masked sequences we expect stage 2 to return (\(M_{1:c}\)). \(c\) can also be viewed as the number of final couplets that we will be expecting stage 3 MLM to mask-predict and return as the final poem.

\begin{equation}
\label{eq:11}
    c = \Bigg \lceil \frac{n^{\prime}}{10} \Bigg \rceil
\end{equation}
Furthermore, we make the heuristic adaptable to the number of keywords it lies between mask tokens to form a sequence. Since members of \(M_{1:c}\) are in couplet-level, we can also specify  the keyword load in each of the right- and left-side hemistichs. Accordingly, we define \(r\) and \(l\) as the number of keywords to be indicated in each masked sequence's right- and left-side hemistichs, respectively. Since each couplet contains two hemistichs only, using Equation \ref{eq:12}, the heuristic calculates the overall number of keywords it needs to form \(c \times 2\) masked hemistichs. 
\begin{equation}
\label{eq:12}
    o = c \times (r +l)
\end{equation}
Comparing the size of the final keywords set \(FK\) and the overall required number of keywords \(o\), we choose to either re-augment or truncate the \(FK\) set to make the two equal when they are in different lengths. Given \(D = |o - n^{\prime}|\), the former process (re-augmentation) refers to a process during which we augment the D least aggregated indices' top frequent keywords using the \(Syn(K_i,1)\) function (expanding the \(FK\) set). In contrast, the latter process (truncation) points to a process during which we remove the D most aggregated indices' least frequent keywords (shrinking the \(FK\) set). Having performed either of the two process, the \(FK\) set now contains \(o\) keywords which are ready to be distributed among mask tokens and form \(c\) sequence.

\subsubsection{Keyword Distribution}
By now, we have collected the final set of keywords representing prose context. In this part, we will distribute them among mask tokens so that proper \(M\) sequences are formed. The properness of these masked sequences heavily rely on how the predicted (keyword) and unpredicted (mask) tokens are aranged next to each other. As \(M\) will be fed into a MLM to predict its masks, we design masked templates in favor of the MLM to produce better text as the final output. Based on our experiments, the adjacency of the keywords inside masked sequences has the highest impact on the quality of final outputs. Hence, we designed this subsection to overcome the keyword adjacency challenge.

Since more than 98\% of all hemistichs available in the Ganjoor Dataset have a word count of less than or equal to 10, we designate a maximum length of 10 tokens for each masked hemistich to tackle our problem. Next, we calculate index-based word frequency distributions \(F^n_{1:10}\) using NLTK \cite{bird2009natural} to measure the presence of the Ganjoor dataset vocabulary in each of the 1 to 10 indices.
\begin{figure*}[!t]
\begin{center}
\includegraphics[width=\linewidth]{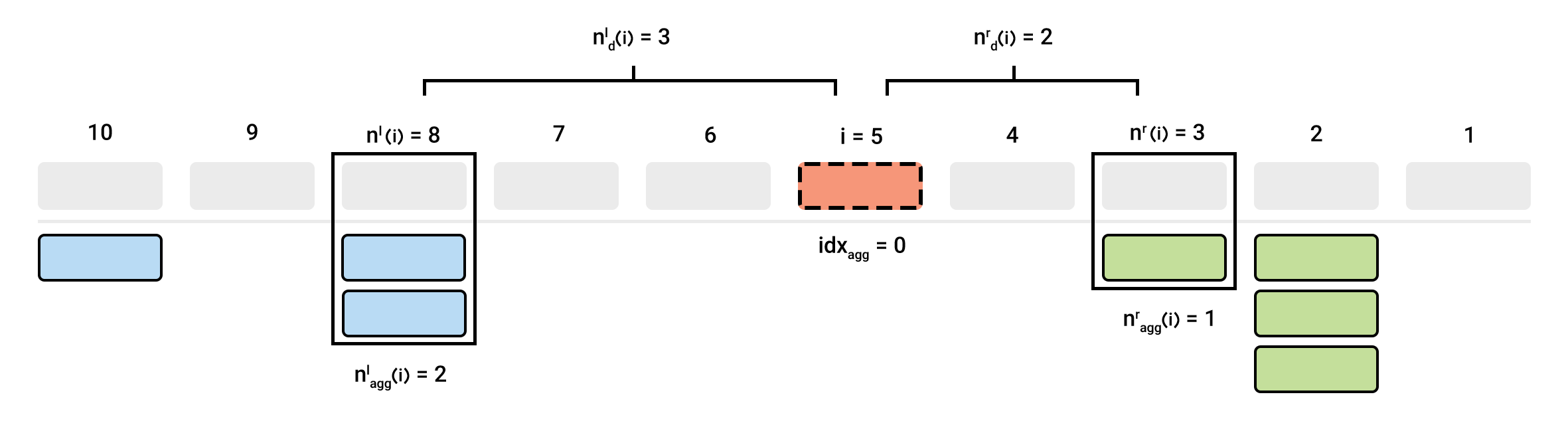}
\end{center}
\caption{An example of nearest right- and left-side neighbours of an arbitrary index (i = 5). Note that indices are ordered from right to left due to the Persian writing style.}
\label{fig:322g}
\end{figure*}
To form logically ordered masked sequences, we intend to distribute keywords among those indices where each keyword a) has previously been most appeared in and b) is logically separated from the rest of the keywords. To do so, we iterate final keywords (\(FK_{1:n^{\prime}}\)) and calculate a series of equations to obtain their final positional indices (\(PI_{1:n^{\prime}}\)). This means that we will take a number of steps to figure out which of the indices between 1 to 10 is most suitable for each of the final keywords. These steps begin with calculating \(PS_{1:10}\) (Equation \ref{eq:3}) per keyword (\(key\)), designed to calculate ten positional scores. Finally, as mentioned in Equation \ref{eq:4}, the hemistich positional index that the keyword will be finally placed in is equal to the argument (index) containing the maximum positional score. 

\begin{equation}
\label{eq:3}
PS_{1:10} = \{ score^{idx}(key, \, i) | 1 \leq i \leq 10 \}
\end{equation}
\begin{equation}
\label{eq:4}
PI_{n^{\prime}} =  argmax(PS_{1:10})
\end{equation}
The most significant calculation here is measuring how likely it would be for the \(key\) to appear in a specific index (\(i\)). In Equation \ref{eq:4}, we define the \(score^{idx}\) scoring function to measure the \(key\)'s belonging to index \(i\). 
\begin{equation}
\label{eq:5}
\small
score^{idx}(key, i) = \frac{\Bigl| \; score^n \Bigl( \: n^r_d(i), \: n^r_{agg}(i), \: 0 \: \Bigl) \; - \; score^n \Bigl( \: n^l_d(i), \: n^l_{agg}(i), \: 1 \: \Bigl) \; \Bigl|}{idx_{agg} + 1} \; \times \; \log_{10}\Bigl(F^n_i(Key)\Bigl)
\end{equation}
Given \(n^r(i)\) and \(n^l(i)\) as the nearest right- and left-side neighbour indices of \(i\), we define \(n^r_d(i)\) and \(n^l_d(i)\) as the distances between \(n^r(i)\) to \(i\) and \(n^l(i)\) to \(i\). Furthermore, \(n^r_{agg}(i)\) and \(n^l_{agg}(i)\) are defined as the number of previously suggested keyword positions (\(PI_{1:n^{\prime}}\)) for the \(n^r(i)\) and \(n^l(i)\) indices, respectively (Figure \ref{fig:322g}). Altogether, the \(score^{idx}\) scoring function deteriorates by the number of previously suggested positions at index \(i\) (\(idx_{agg}\)) and improves with two factors, where one is the frequency of \(key\) in index \(i\) (indicated as \(F^n_i(key)\)) and the other is the absolute difference between the right-side (\(score^n(..., \, ..., 0)\)) and the left-side (\(score^n(..., \, ..., 1)\)) neighbour scores. The intuition here is that the less aggregated a side is, the more appealing that index (\(i\)) is for \(key\) to be placed in. 
\begin{equation}
\label{eq:6}
\small
score^n(x, \, y, \, z) = \frac{x+1}{\Bigl(score^{agg}(\: i, \: z \: ) \, + \, 1 \: \Bigl) \; \times \; \Bigl(y \, + \, 1 \Bigl)}
\end{equation}
To score a neighbour side, we define \(score^n\) (Equation \ref{eq:6}) where it improves by the further its nearest neighbour (i.e. \(n^r(i)\) or \(n^l(i)\)) is and degrades by the more keyword aggregated it is. We use Equation \ref{eq:8} to define \(score^{agg}\) as a measurement of how aggregated the right (\(j=0\)) or left (\(j=1\)) neighbour side of \(i\) is.
\begin{equation}
\label{eq:8}
 score^{agg}(x, y) = \\
 \begin{cases} 
    \: \bigg| \, \{ \: j \: | \: 1 \leq j \leq n^{\prime} \: , \: PI_{j} < x \: \} \, \bigg| & y = 0 \\ \\
    \: \bigg| \, \{ \: j \: | \: 1 \leq j \leq n^{\prime} \: , \: PI_{j} > x \: \} \, \bigg| & y = 1 
  \end{cases}
\end{equation}
Having iterated all \(FK_{1:n^{\prime}}\) elements, we obtain the full \(PI_{1:n^{\prime}}\) set, which contains indices that each \(FK\) set member would best fit in. Lastly, we create the \(IK_{1:10}\) set where it has a list of all suggested keywords for each of the 10 indices. The more uniformly distributed the keywords are among the indices, the more likely it is to form logically ordered associated MKSs.

\subsubsection{Word Associations Graph}
To make the heuristic knowledgeable of poetic word associations, we seek to capture a pattern among available ones. However, associated words in Persian poetry cannot be easily extracted. Figures of Speech have long played prominent roles in the significance of Persian Poetry. Based on aesthetic values, they add non-literary adjustments, which may, in certain scenarios, make the poem undergo meaning divergence. Hence, defining a valid association as "a set of contextually similar words" is inadequate. Instead, we use the graph data structure to model associations indicated in our dataset's poems to capture unbiased knowledge. This way, we can simultaneously investigates both the \textit{spatial attitudes} \cite{seanatt} and the \textit{context of agreement} among associated words.
 
Using the NetworkX \cite{SciPyProceedings_11} python package, we construct a graph with \(V\) vertices, each representing a word in the Ganjoor dataset's vocabulary. Then we iterate couplets to extract unique word combinations of two and draw an edge between them at their earliest co-occurrence. Next, given \(v_1\) and \(v_2\) as two connected vertices and \(v^i_1\) and \(v^i_2\) as their indices (ranging from 1 to 20), we define edge attribute \(E(v_1, v_2)\) using Equation \ref{eq:13}, which scores the two vertices co-occurrence by their spatial distance.
\begin{equation}
\label{eq:13}
    E(v_1, v_2) = 1 - \frac{\big| i - j \big|}{20}
\end{equation}

While some figures of speech, including paradox, make semantic similarity an invalid measure of how poetically two words are associated in Persian poetry, several other figures of speech, such as synonym, heavily rely on it. Consequently, it would be beneficial to encourage contextually similar word pairs and penalize contextually dissimilar ones. To do so, we train a Word2Vec model \cite{mikolov2013efficient} on the Ganjoor dataset with a window size of 5 to map words onto 64-dimensional spaces.

Since the two \(v_1\) and \(v_2\) words can co-occure multiple times, we calculate \(E(v_1, v_2)\) each time. Moreover, we define \(\overline{E(v_1, v_2)}\) as the mean of all previously-calculated edge attributes and \(sim\) as \(v_1\) and \(v_2\) word vectors' cosine similarity. Finally, We define their final association score as the arithmetic mean of \(\overline{E(v_1, v_2)}\) and \(sim\). Final association scores will further be obtained from the graph to identify how much poetically associated word pairs are. The more similar and spatially close two words are, the more likely they are to be poetically associated.

\subsubsection{Partitioning}
Thus far, we have coped with Presence and Association-related challenges, which form components of a standard poem. However, the adjacency of keywords in masked sequences effectively contributes to the quality of final poems as well. This step is designed to extract a set of masked sequences with fairly distributed keywords in-between mask tokens such that are A) adequate in number, B) literary/contextually associated, and C) logically separated among mask tokens. After adding a mask token to each \(IK_{1:10}\) index-based keyword sets, we score possible occurrences using Equation \ref{eq:14} to satisfy these criteria.
\begin{equation}
\label{eq:14}
    score^{p}(d, g) = sigmoid\big(\log_{2} (d) \times g \big)
\end{equation}
To score each candidate combination, we step through every two adjacent keywords (\(v_1\), and \(v_2\)) to calculate their pair-wise spatial differences and final graph-based association scores. Given \(d\) as the sum of all pair-wise spatial differences and \(g\) as the mean of all pair-wise association scores, \(score^{p}(d, g)\) is calculated for every possible masked candidate. Finally, a total of \(c \times 2\) top masked hemistichs are selected as final candidates and zipped index-wise to form couple-level masked sequences. 

\subsubsection{Poetry Format  Adaptation}
Persian Poems can be categorized into different formats such as \textit{Ghazal}, \textit{Masnavi}, \textit{Dobeiti}, or \textit{Ghet'e} where each has its unique way of locating rhymes inside couplets. As these locations are pre-determined in each format, the challenge becomes finding words that rhyme with each other so that we can place them in those specific locations. Thus, We gathered a list of words that rhyme together and used them to make poems adaptable to user's format of interest. To select rhymes for masked sequences, we chose an initial rhyme and further explored other words that rhyme with it. It is noteworthy that this selection is highly based on how well-associated a rhyme candidate is with the masked template. More specifically, we chose the candidate with the highest mean of pair-wise association scores concerning previously-selected rhymes and indicated keywords inside the masked sequence. This process enabled the heuristic to select contextually-relevant rhymes per each masked sequence and take the final step towards better-formed \(M_{1:c}\) set members.

\subsection{Stage 3: Final Translation}
Having generated a set of masked templates containing poetically-arranged keywords, we define stage 3 as a function that maps masked sequences to final poems (Equation \ref{eq:15}). 
\begin{equation}
\label{eq:15}
    Stage \: 3:M_{1:c} \rightarrow Y_{1:c}
\end{equation}
To do so, we use state-of-the-art MLMs to mask-predict the template and build a full couplet by filling in the blanks. As discussed previously, instead of using encoder-decoder models and let them be fully in charge of translating prose to standard Persian poems, we asked it to solely generate a contextually-rich initial translation of the input prose and let stages 2 and 3 be in charge of poetry generation. Besides, since we have a relatively larger amount of unlabeled monolingual poems, we pretrained different variations of BERT on the Ganjoor Dataset to let it update its general knowledge of the Persian to learn latent ancient poetry features. In terms of masked language modelling, while available BERT models used a 15\% probability of masking input tokens during their initial training phase, we set BERT models to randomly mask 40\% during the pretraining phase to expect more of the model by giving it less and expose it to similar context to what it will see during the poetry generation stage. We pretrained available Persian MLMs such as BERT, RoBERTa, ALBERT, and DistilBERT on the Ganjoor Dataset. In terms of length, we experimented with both hemistich-level and Couplet-level pretraining, among which the latter performed significantly better in predicting a contextually rich set of masked sequences.

Since MLMs predict a probability distribution over each word to predict a mask, we need a strategy to decode the final poem by selecting between a set of candidates per mask. We approach this using a beam search decoder with a customised heuristic inspired by \citet{insp1} and \citet{insp2}.
\begin{equation}
\label{eq:16}
    score^b(g_4, g_3, g_2) = \frac{(4 \times g_4) + (3 \times g_3) + (2 \times g_2)}{10}
\end{equation}

Given \(g_4\), \(g_3\), and \(g_2\) as the 4-gram, trigram, and bigram of each mask-predicted candidate word and its previously-indicated words, the beam search decoder constantly ranks the \(k\) (Beam Depth) most probable sequences using Equation \ref{eq:16}. Predicted sequences using the beam search algorithm are our approach's final decoded poems (\(Y_{1:c}\)).

\begin{table*}[t]
\centering
\caption{Comparison between different approaches (SA = Semantic Affinity, PL = Perplexity, B = BLEU, R = ROUGE)}
\label{tab:mainResult}
\begin{tabular*}{\textwidth}{c @{\extracolsep{\fill}} llcccccccc}
\toprule
& \textbf{Stage} & \textbf{Approach} & \textbf{SA} & \textbf{PL} & \textbf{B-1} & \textbf{B-2} & \textbf{B-3} & \textbf{R-1} & \textbf{R-2} & \textbf{R-L}\\
\midrule
1 & IT & Seq2Seq & 0.15 & 55.14 & 0.1721 & 0.0576 & 0.1930 & 0.1010 & 0.0001 & 0.1326\\
2 & IT & Seq2Seq + Attention & 0.36 & 29.96 & 0.2784 & 0.2011 & 0.1508 & 0.3961 & 0.2508 & 0.4398 \\
3 & IT & Seq2Seq + Transformer & 0.57 & \textbf{12.18} & 0.2777 & 0.2617 & 0.1675 & 0.6676 & 0.4839 & 0.6643 \\
\hline
4 & FT &  \(h\) + BERT (h) & 0.60 & 66.63 & 0.8435 & 0.3323 & 0.2092 & 0.8776 & 0.3725 & 0.8540\\
5 & FT &  \(h\) + DistilBERT (h) & 0.61 & 69.14 & 0.8507 & 0.3463 & 0.2009 & 0.8792 & \textbf{0.3882} & 0.8587 \\
6 & FT &  \(h\) + ALBERT (h) & 0.59 & 69.39 & 0.8326 & 0.3004 & \textbf{0.2244} & 0.8580 & 0.3459 & 0.8334 \\
7 & FT &  \(h\) + RoBERTa (h) & 0.65 & 61.86 & \textbf{0.8700} & \textbf{0.3512} & 0.2010 & \textbf{0.8976} & 0.3857 & \textbf{0.8769} \\
\hline
8 & FT &  \(h\) + BERT (c) & \textbf{0.72} & 60.89 & 0.8386 & 0.2991 & 0.2032 & 0.8738 & 0.3376 & 0.8504 \\
9 & FT &  \(h\) + DistilBERT (c) & 0.67 & 82.64 & 0.8444 & 0.3040 & 0.2102 & 0.8854 & 0.3374 & 0.8637 \\
10 & FT &  \(h\) + ALBERT (c) & 0.69 & \textbf{57.37} & 0.8385 & 0.2585 & 0.2224 & 0.8721 & 0.3002 & 0.8473 \\
11 & FT &  \(h\) + RoBERTa (c) & 0.67 & 68.98 & 0.8509 & 0.3150 & 0.2132 & 0.8894 & 0.3610 & 0.8696 \\
\bottomrule
\end{tabular*}
\end{table*}

\begin{table*}[t]
\caption{Human Annotations' Comparision (F = Fluency, C = Coherence, M = Meaningfulness, PO = Poeticness, T = Translation, P = Professional, NP = Non-professional)}
\label{tab:mainHuman}
\centering
\begin{tabular*}{\textwidth}{c @{\extracolsep{\fill}} llccccccccc}
\toprule
& \textbf{Approach} & \textbf{F-NP} & \textbf{C-NP} & \textbf{M-NP} & \textbf{PO-NP} & \textbf{T-NP} & \textbf{F-P} & \textbf{C-P} & \textbf{M-P} & \textbf{PO-P} & \textbf{T-P}\\
\midrule
% & S & 3.9985 & 3.8549 & 3.7668 & 3.8379 & 3.6243 & 3.055 & 3.0952 & 2.9525 & 2.875 & 4.0825 & 1.9825\\
& Seq2Seq + Tra... & 2.4842 & 2.3742 & 2.2935 & 2.3812 & 2.1938 & 1.5528 & 1.5880 & 1.4774 & 1.4136 & \textbf{2.6323}\\
& \(h\) + RoBERTa (c) & 3.6400 & \textbf{3.6625} & \textbf{3.6675} & \textbf{3.8400} & \textbf{3.7100} & 2.4900 & 2.8275 & 2.4575 & \textbf{2.6375} & 2.3225\\
& \(h\) +  RoBERTa (h) & \textbf{3.8564} & 3.4309 & 3.3174 & 3.2932 & 3.2643 & \textbf{2.6197} & \textbf{3.9575} & \textbf{3.7050} & 2.4300 & 2.3550\\
\toprule
\end{tabular*}
\end{table*}

\section{Experiments \& Results}
Our curated parallel corpus contains 5,191 pairs of prose and poems. First, we left 1,720 pairs for evaluation and augmented the remaining (train set) up to 28,820 pairs using the \textit{Word Substitution} technique to train encoder-decoder (stage 1) models for 13 epochs and with an embedding dimension of 256. The Monolingual pretraining phase of MLMs (stage 3) was carried out using 321,581 extracted couplets from the Ganjoor Dataset\footnote{www.ganjoor.net} (train=289,422 test=32,159). We pretrained BERT, DistilBERT \cite{DistilBERT}, ALBERT \cite{ALBERT}, and RoBERTa \cite{RoBERTa} for 5 epochs using experimentally determined values of 0.40 and 0.15 masked token ratios in couplet-level (c) and hemistich-level (h) MLMs. To augment keywords in the heuristic (stage 2), we collected a dataset of synonyms and antonyms for each word in the vocabulary, totaling 4,756 words with an average synonym and antonym count of 7.12 and 2.34, respectively. 

We used both automatic and human evaluation metrics to evaluate our experimental approaches. BLEU \cite{articlePap} and ROUGE \cite{lin-2004-rouge} are among the most conventional metrics in evaluating text generation tasks. As part of human evaluation, two groups of 600 and 100 random poems were assessed by both poetry professionals and non-professionals, respectively. Moreover, we used Perplexity (PL) to measure the generalization performance of language models. We also used another metric to measure the semantic affinity between generated and ground-truth poems through thematically classifying them. To do so, we gathered a dataset of 1,319 couplets (augmented up to 2,567) and thematically labelled them as Divine (\textit{Elahi}), Ethical (\textit{Akhlaghi}), Amorous (\textit{Asheghaneh}), or Philosophical (\textit{Erfani}). Generally, this metric depicts what percentage of translated and ground-truth poems describe the same concept, enabling a high-level understanding of how semantically similar the two are.

As indicated in Table \ref{tab:mainResult}, we experimented with different Initial Translation (IT) and Final Translation (FT) approaches. According to the first three rows, seq2seq architectures suffer in low-resource settings where scarce data prevents them from generalizing, which leads to poor semantic affinity, BLEU, and ROUGE scores. In contrast, heuristically-aided FT models (rows 4 to 11) have overcome the data scarcity challenge by exploiting monolingual data and quality-control heuristics. A manifestation of this is witnessed when transitioning from IT to FT approaches, where scores have improved significantly. Generally, the combination of H and ROBERTA (rows 7 and 11) has yielded relatively higher scores among BLEU and ROUGE metrics in both couplet and hemistich levels. While couplet-level approaches such as \(h\) + BERT and \(h\) + ALBERT have achieved better results in terms of semantic affinity and perplexity. However, such unique tasks cannot be simply evaluated by automatic metrics, and more intelligence is needed for this purpose \cite{han2018machine}.

Following the work of \citet{van-de-cruys-2020-automatic}, we asked a group of professionals (P) and non-professionals (NP) to annotate leading approaches' generated poems by their \textit{Fluency} (F), \textit{Coherence} (C), \textit{Meaningfulness} (M), \textit{Poeticness} (PO), and \textit{Translation Quality} (T) from 1 to 5. The professionals were further asked to rate the generated poems on a scale of 1 to 3 based on how much they replicated previous literature and how unoriginal they were. As shown in Table \ref{tab:mainHuman}, leading MLMs, performed considerably better than transformer-based Encoder-Decoder models in most metrics. 

\bgroup
\def\arraystretch{1}
\small
\begin{longtable}{p{0.1\textwidth}|p{0.85\textwidth}}
  \caption{Random prose and their respective processed text in each step. An English equivalent of the initial input and the final outcome is provided among all samples. (G = Ground truth)} \label{tab:ourtabel_1} \\
    \hline
     Type &  \hspace{12.3cm} Text \\ 
    %  \hline
    %  \hline
    % \multicolumn{2}{c}{\textbf{Sample 1}} \\
    %  \hline
    % \multirow{ 2}{*}{}Prose & 
    % \begin{minipage}{\textwidth}
    % \vspace{1mm}
    % \raggedleft
    % \includegraphics[scale=0.35]{Samples/1_1.JPG}
    % \hspace{2.08cm}
    % \vspace{1mm}
    % \end{minipage} \\
    % \hline
    %  Poem (G) & \begin{minipage}{\textwidth}
    % \vspace{1mm}
    % \raggedleft
    % \includegraphics[scale=0.35]{Samples/1_2.JPG}
    % \hspace{2.08cm}
    % \vspace{1mm}
    % \end{minipage} \\
    %  \hline
    %  IT & \begin{minipage}{\textwidth}
    % \vspace{1mm}
    % \raggedleft
    % \includegraphics[scale=0.35]{Samples/1_3.JPG}
    % \hspace{2.08cm}
    % \vspace{1mm}
    % \end{minipage} \\
    %  \hline
    %  Heuristic & \begin{minipage}{\textwidth}
    % \vspace{1mm}
    % \raggedleft
    % \includegraphics[scale=0.35]{Samples/1_4.JPG}
    % \hspace{2.08cm}
    % \vspace{1mm}
    % \end{minipage} \\
    %  \hline
    %  FT & \begin{minipage}{\textwidth}
    % \vspace{1mm}
    % \raggedleft
    % \includegraphics[scale=0.35]{Samples/1_5.JPG}
    % \hspace{2.08cm}
    % \vspace{1mm}
    % \end{minipage} \\
     \hline
     \hline
    \multicolumn{2}{c}{\textbf{Sample 1}} \\
     \hline
     \multirow{ 2}{*}{}Prose & \begin{minipage}{\textwidth}
    \vspace{1mm}
    \raggedleft
    \includegraphics[scale=0.35]{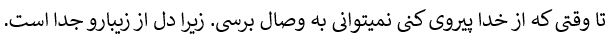} 
    \hspace{2.08cm}
    \vspace{1mm}
    \end{minipage} 
    \newline
    It is only by following the almighty that you can achieve your ultimate destiny. Because the heart is separate from the beauty.\\
    \hline
     Poem (G) & \begin{minipage}{\textwidth}
    \vspace{1mm}
    \raggedleft
    \includegraphics[scale=0.35]{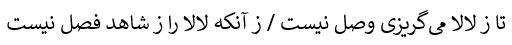}
    \hspace{2.08cm}
    \vspace{1mm}
    \end{minipage} \\
     \hline
     IT & \begin{minipage}{\textwidth}
    \vspace{1mm}
    \raggedleft
    \includegraphics[scale=0.35]{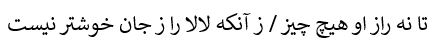}
    \hspace{2.08cm}
    \vspace{1mm}
    \end{minipage} \\
     \hline
     Heuristic & \begin{minipage}{\textwidth}
    \vspace{1mm}
    \raggedleft
    \includegraphics[scale=0.35]{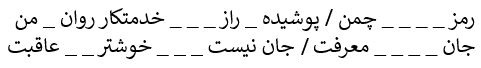}
    \hspace{2.08cm}
    \vspace{1mm}
    \end{minipage} \\
     \hline
     FT & \begin{minipage}{\textwidth}
    \vspace{1mm}
    \raggedleft
    \includegraphics[scale=0.35]{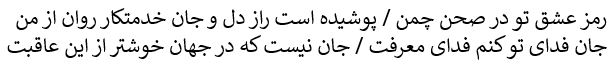}
    \hspace{2.08cm}
    \vspace{1mm}
    \end{minipage} 
    \newline
    Your mark can be seen in creations that are occult to my sight. The divine will is my ultimate future, and I will sacrifice my life for it. \\
    %  \hline
    %  \hline
    % \multicolumn{2}{c}{\textbf{Sample 3}} \\
    %  \hline
    %  \multirow{ 2}{*}{}Prose & \begin{minipage}{\textwidth}
    % \vspace{1mm}
    % \raggedleft
    % \includegraphics[scale=0.35]{Samples/3_1.JPG}
    % \hspace{2.08cm}
    % \vspace{1mm}
    % \end{minipage} \\
    % \hline
    %  Poem (G) & \begin{minipage}{\textwidth}
    % \vspace{1mm}
    % \raggedleft
    % \includegraphics[scale=0.35]{Samples/3_2.JPG}
    % \hspace{2.08cm}
    % \vspace{1mm}
    % \end{minipage} \\
    %  \hline
    %  IT & \begin{minipage}{\textwidth}
    % \vspace{1mm}
    % \raggedleft
    % \includegraphics[scale=0.35]{Samples/3_3.JPG}
    % \hspace{2.08cm}
    % \vspace{1mm}
    % \end{minipage} \\
    %  \hline
    %  Heuristic & \begin{minipage}{\textwidth}
    % \vspace{1mm}
    % \raggedleft
    % \includegraphics[scale=0.35]{Samples/3_4.JPG}
    % \hspace{2.08cm}
    % \vspace{1mm}
    % \end{minipage} \\
    %  \hline
    %  FT & \begin{minipage}{\textwidth}
    % \vspace{1mm}
    % \raggedleft
    % \includegraphics[scale=0.35]{Samples/3_5.JPG}
    % \hspace{2.08cm}
    % \vspace{1mm}
    % \end{minipage} \\
     \hline
     \hline
    \multicolumn{2}{c}{\textbf{Sample 2}} \\
     \hline
     \multirow{ 2}{*}{}Prose & \begin{minipage}{\textwidth}
    \vspace{1mm}
    \raggedleft
    \includegraphics[scale=0.35]{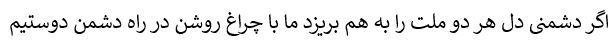}
    \hspace{2.08cm}
    \vspace{1mm}
    \end{minipage}
    \newline
    We will become enlightened friends with the enemy if enmity breaks both nations' hearts. \\
    \hline
     Poem (G) & \begin{minipage}{\textwidth}
    \vspace{1mm}
    \raggedleft
    \includegraphics[scale=0.35]{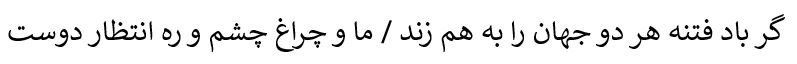}
    \hspace{2.08cm}
    \vspace{1mm}
    \end{minipage} \\
     \hline
     IT & \begin{minipage}{\textwidth}
    \vspace{1mm}
    \raggedleft
    \includegraphics[scale=0.35]{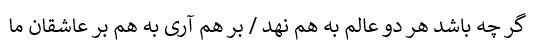}
    \hspace{2.08cm}
    \vspace{1mm}
    \end{minipage} \\
     \hline
     Heuristic & \begin{minipage}{\textwidth}
    \vspace{1mm}
    \raggedleft
    \includegraphics[scale=0.35]{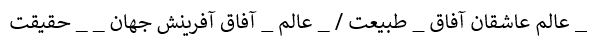}
    \hspace{2.08cm}
    \vspace{1mm}
    \end{minipage} \\
     \hline
     FT & \begin{minipage}{\textwidth}
    \vspace{1mm}
    \raggedleft
    \includegraphics[scale=0.35]{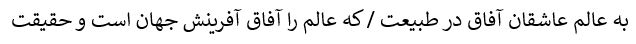}
    \hspace{2.08cm}
    \vspace{1mm}
    \end{minipage} 
    \newline
    Since the wisest see in the universe creation and truth, the world of lovers is better. \\
     \hline
     \hline
    \multicolumn{2}{c}{\textbf{Sample 3}} \\
     \hline
     \multirow{ 2}{*}{}Prose & \begin{minipage}{\textwidth}
    \vspace{1mm}
    \raggedleft
    \includegraphics[scale=0.35]{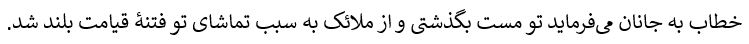}
    \hspace{2.08cm}
    \vspace{1mm}
    \end{minipage}
    \newline
    Since the angels saw you passing by, intoxicated, the tumult of resurrection rose.  \\
    \hline
     Poem (G) & \begin{minipage}{\textwidth}
    \vspace{1mm}
    \raggedleft
    \includegraphics[scale=0.35]{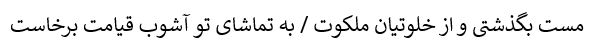}
    \hspace{2.08cm}
    \vspace{1mm}
    \end{minipage} \\
     \hline
     IT & \begin{minipage}{\textwidth}
    \vspace{1mm}
    \raggedleft
    \includegraphics[scale=0.35]{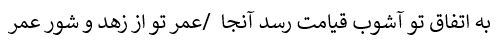}
    \hspace{2.08cm}
    \vspace{1mm}
    \end{minipage} \\
     \hline
     Heuristic & \begin{minipage}{\textwidth}
    \vspace{1mm}
    \raggedleft
    \includegraphics[scale=0.35]{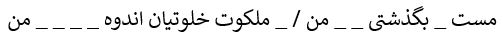}
    \hspace{2.08cm}
    \vspace{1mm}
    \end{minipage} \\
     \hline
     FT & \begin{minipage}{\textwidth}
    \vspace{1mm}
    \raggedleft
    \includegraphics[scale=0.35]{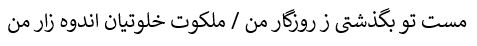}
    \hspace{2.08cm}
    \vspace{1mm}
    \end{minipage}
    \newline
    You passed by my life, intoxicated, and the world of angles is where I grieve. \\
     \hline
     \hline
    \multicolumn{2}{c}{\textbf{Sample 4}} \\
     \hline
     \multirow{ 2}{*}{}Prose & \begin{minipage}{\textwidth}
    \vspace{1mm}
    \raggedleft
    \includegraphics[scale=0.35]{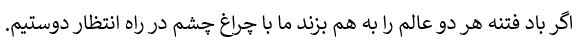}
    \hspace{2.08cm}
    \vspace{1mm}
    \end{minipage} 
    \newline
    Even if sedition revolts the two worlds, we will be waiting for the one. \\
    \hline
     Poem (G) & \begin{minipage}{\textwidth}
    \vspace{1mm}
    \raggedleft
    \includegraphics[scale=0.35]{Samples/4_2.JPG}
    \hspace{2.08cm}
    \vspace{1mm}
    \end{minipage} \\
     \hline
     IT & \begin{minipage}{\textwidth}
    \vspace{1mm}
    \raggedleft
    \includegraphics[scale=0.35]{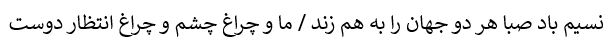}
    \hspace{2.08cm}
    \vspace{1mm}
    \end{minipage} \\
     \hline
     Heuristic & \begin{minipage}{\textwidth}
    \vspace{1mm}
    \raggedleft
    \includegraphics[scale=0.35]{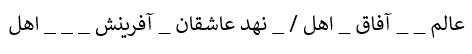}
    \hspace{2.08cm}
    \vspace{1mm}
    \end{minipage} \\
     \hline
     FT & \begin{minipage}{\textwidth}
    \vspace{1mm}
    \raggedleft
    \includegraphics[scale=0.35]{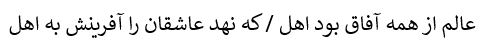}
    \hspace{2.08cm}
    \vspace{1mm}
    \end{minipage}
    \newline
    The universe belongs to all, which lets the creation make lovers descent. \\
    %  \hline
    %  \hline
    % \multicolumn{2}{c}{\textbf{Sample 5}} \\
    %  \hline
    %  \multirow{ 2}{*}{}Prose & \begin{minipage}{\textwidth}
    % \vspace{1mm}
    % \raggedleft
    % \includegraphics[scale=0.35]{Samples/7_1.JPG}
    % \hspace{2.08cm}
    % \vspace{1mm}
    % \end{minipage} \\
    % \hline
    %  Poem (G) & \begin{minipage}{\textwidth}
    % \vspace{1mm}
    % \raggedleft
    % \includegraphics[scale=0.35]{Samples/7_2.JPG}
    % \hspace{2.08cm}
    % \vspace{1mm}
    % \end{minipage} \\
    %  \hline
    %  IT & \begin{minipage}{\textwidth}
    % \vspace{1mm}
    % \raggedleft
    % \includegraphics[scale=0.35]{Samples/7_3.JPG}
    % \hspace{2.08cm}
    % \vspace{1mm}
    % \end{minipage} \\
    %  \hline
    %  Heuristic & \begin{minipage}{\textwidth}
    % \vspace{1mm}
    % \raggedleft
    % \includegraphics[scale=0.35]{Samples/7_4.JPG}
    % \hspace{2.08cm}
    % \vspace{1mm}
    % \end{minipage} \\
    %  \hline
    %  FT & \begin{minipage}{\textwidth}
    % \vspace{1mm}
    % \raggedleft
    % \includegraphics[scale=0.35]{Samples/7_5.JPG}
    % \hspace{2.08cm}
    % \vspace{1mm}
    % \end{minipage} \\
     \hline
     \hline
    \multicolumn{2}{c}{\textbf{Sample 5}} \\
     \hline
     \multirow{ 2}{*}{}Prose & \begin{minipage}{\textwidth}
    \vspace{1mm}
    \raggedleft
    \includegraphics[scale=0.35]{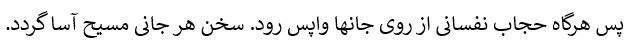}
    \hspace{2.08cm}
    \vspace{1mm}
    \end{minipage} 
    \newline
    By pulling back the curtains, the words of each individual will be curative. \\
    \hline
     Poem (G) & \begin{minipage}{\textwidth}
    \vspace{1mm}
    \raggedleft
    \includegraphics[scale=0.35]{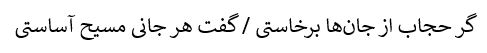}
    \hspace{2.08cm}
    \vspace{1mm}
    \end{minipage} \\
     \hline
     IT & \begin{minipage}{\textwidth}
    \vspace{1mm}
    \raggedleft
    \includegraphics[scale=0.35]{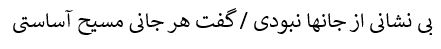}
    \hspace{2.08cm}
    \vspace{1mm}
    \end{minipage} \\
     \hline
     Heuristic & \begin{minipage}{\textwidth}
    \vspace{1mm}
    \raggedleft
    \includegraphics[scale=0.35]{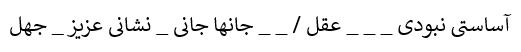}
    \hspace{2.08cm}
    \vspace{1mm}
    \end{minipage} \\
     \hline
     FT & \begin{minipage}{\textwidth}
    \vspace{1mm}
    \raggedleft
    \includegraphics[scale=0.35]{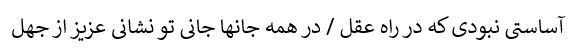}
    \hspace{2.08cm}
    \vspace{1mm}
    \end{minipage}
    \newline
    With regard to wisdom, you were not demure, and among all individuals, you are ignorant. \\
     \hline
     \hline
    \multicolumn{2}{c}{\textbf{Sample 6}} \\
     \hline
     \multirow{ 2}{*}{}Prose & \begin{minipage}{\textwidth}
    \vspace{1mm}
    \raggedleft
    \includegraphics[scale=0.35]{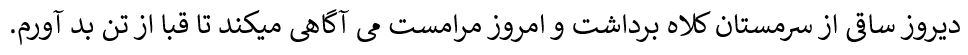}
    \hspace{2.08cm}
    \vspace{1mm}
    \end{minipage} 
    \newline
    Yesterday, the cupbearer took off the hats of the intoxicated. Now, he intoxicates the believers to defrock them. \\
    \hline
     Poem (G) & \begin{minipage}{\textwidth}
    \vspace{1mm}
    \raggedleft
    \includegraphics[scale=0.35]{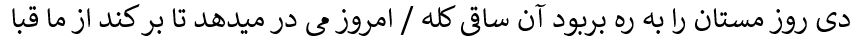}
    \hspace{2.08cm}
    \vspace{1mm}
    \end{minipage} \\
     \hline
     IT & \begin{minipage}{\textwidth}
    \vspace{1mm}
    \raggedleft
    \includegraphics[scale=0.35]{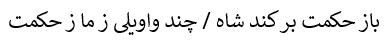}
    \hspace{2.08cm}
    \vspace{1mm}
    \end{minipage} \\
     \hline
     Heuristic & \begin{minipage}{\textwidth}
    \vspace{1mm}
    \raggedleft
    \includegraphics[scale=0.35]{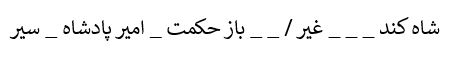}
    \hspace{2.08cm}
    \vspace{1mm}
    \end{minipage} \\
     \hline
     FT & \begin{minipage}{\textwidth}
    \vspace{1mm}
    \raggedleft
    \includegraphics[scale=0.35]{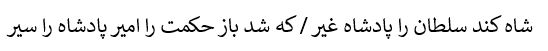}
    \hspace{2.08cm}
    \vspace{1mm}
    \end{minipage}
    \newline
    The king elevates the sultan, which opens up wisdom to its sovereign. \\
     \hline
     \hline
    \multicolumn{2}{c}{\textbf{Sample 7}} \\
     \hline
     \multirow{ 2}{*}{}Prose & \begin{minipage}{\textwidth}
    \vspace{1mm}
    \raggedleft
    \includegraphics[scale=0.35]{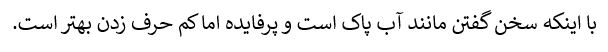}
    \hspace{2.08cm}
    \vspace{1mm}
    \end{minipage}
    \newline
    Although speaking is like pure water and lucrative, but less of it is better. \\
    \hline
     Poem (G) & \begin{minipage}{\textwidth}
    \vspace{1mm}
    \raggedleft
    \includegraphics[scale=0.35]{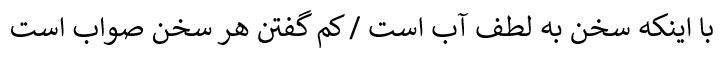}
    \hspace{2.08cm}
    \vspace{1mm}
    \end{minipage} \\
     \hline
     IT & \begin{minipage}{\textwidth}
    \vspace{1mm}
    \raggedleft
    \includegraphics[scale=0.35]{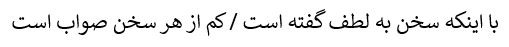}
    \hspace{2.08cm}
    \vspace{1mm}
    \end{minipage} \\
     \hline
     Heuristic & \begin{minipage}{\textwidth}
    \vspace{1mm}
    \raggedleft
    \includegraphics[scale=0.35]{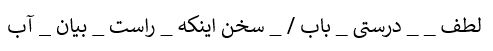}
    \hspace{2.08cm}
    \vspace{1mm}
    \end{minipage} \\
     \hline
     FT & \begin{minipage}{\textwidth}
    \vspace{1mm}
    \raggedleft
    \includegraphics[scale=0.35]{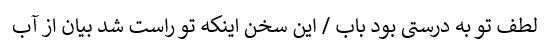}
    \hspace{2.08cm}
    \vspace{1mm}
    \end{minipage} 
    \newline
    Your grace is fairly popularized and your words are true like water. \\
    %  \hline
    %  \hline
    % \multicolumn{2}{c}{\textbf{Sample 11}} \\
    %  \hline
    %  \multirow{ 2}{*}{}Prose & \begin{minipage}{\textwidth}
    % \vspace{1mm}
    % \raggedleft
    % \includegraphics[scale=0.35]{Samples/11_1.JPG}
    % \hspace{2.08cm}
    % \vspace{1mm}
    % \end{minipage} \\
    % \hline
    %  Poem (G) & \begin{minipage}{\textwidth}
    % \vspace{1mm}
    % \raggedleft
    % \includegraphics[scale=0.35]{Samples/11_2.JPG}
    % \hspace{2.08cm}
    % \vspace{1mm}
    % \end{minipage} \\
    %  \hline
    %  IT & \begin{minipage}{\textwidth}
    % \vspace{1mm}
    % \raggedleft
    % \includegraphics[scale=0.35]{Samples/11_3.JPG}
    % \hspace{2.08cm}
    % \vspace{1mm}
    % \end{minipage} \\
    %  \hline
    %  Heuristic & \begin{minipage}{\textwidth}
    % \vspace{1mm}
    % \raggedleft
    % \includegraphics[scale=0.35]{Samples/11_4.JPG}
    % \hspace{2.08cm}
    % \vspace{1mm}
    % \end{minipage} \\
    %  \hline
    %  FT & \begin{minipage}{\textwidth}
    % \vspace{1mm}
    % \raggedleft
    % \includegraphics[scale=0.35]{Samples/11_5.JPG}
    % \hspace{2.08cm}
    % \vspace{1mm}
    % \end{minipage} \\
     \hline
     \hline
    \multicolumn{2}{c}{\textbf{Sample 8}} \\
     \hline
     \multirow{ 2}{*}{}Prose & \begin{minipage}{\textwidth}
    \vspace{1mm}
    \raggedleft
    \includegraphics[scale=0.35]{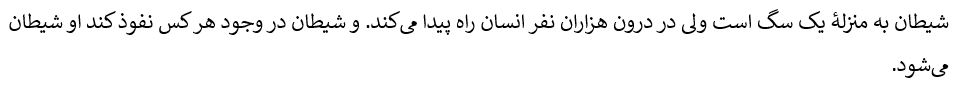}
    \hspace{2.08cm}
    \vspace{1mm}
    \end{minipage} 
    \newline
    The devil seeks to govern thousands of people, and anyone who is ruled by him also becomes the devil. \\
    \hline
     Poem (G) & \begin{minipage}{\textwidth}
    \vspace{1mm}
    \raggedleft
    \includegraphics[scale=0.35]{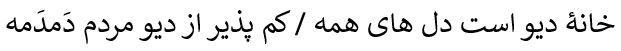}
    \hspace{2.08cm}
    \vspace{1mm}
    \end{minipage} \\
     \hline
     IT & \begin{minipage}{\textwidth}
    \vspace{1mm}
    \raggedleft
    \includegraphics[scale=0.35]{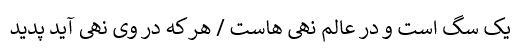}
    \hspace{2.08cm}
    \vspace{1mm}
    \end{minipage} \\
     \hline
     Heuristic & \begin{minipage}{\textwidth}
    \vspace{1mm}
    \raggedleft
    \includegraphics[scale=0.35]{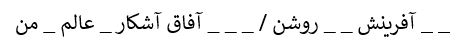}
    \hspace{2.08cm}
    \vspace{1mm}
    \end{minipage} \\
     \hline
     FT & \begin{minipage}{\textwidth}
    \vspace{1mm}
    \raggedleft
    \includegraphics[scale=0.35]{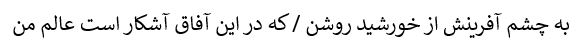}
    \hspace{2.08cm}
    \vspace{1mm}
    \end{minipage} 
    \newline
    A look at creation reveals the almighty power and the horizon reveals my true world. \\
     \hline
     \hline
    \multicolumn{2}{c}{\textbf{Sample 9}} \\
     \hline
     \multirow{ 2}{*}{}Prose & \begin{minipage}{\textwidth}
    \vspace{1mm}
    \raggedleft
    \includegraphics[scale=0.35]{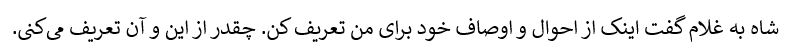}
    \hspace{2.08cm}
    \vspace{1mm}
    \end{minipage} 
    \newline
    The king told the servant to stop defining others and define himself. \\
    \hline
     Poem (G) & \begin{minipage}{\textwidth}
    \vspace{1mm}
    \raggedleft
    \includegraphics[scale=0.35]{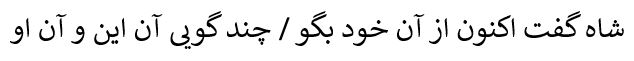}
    \hspace{2.08cm}
    \vspace{1mm}
    \end{minipage} \\
     \hline
     IT & \begin{minipage}{\textwidth}
    \vspace{1mm}
    \raggedleft
    \includegraphics[scale=0.35]{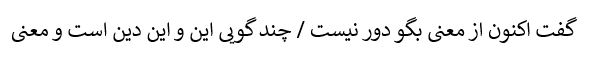}
    \hspace{2.08cm}
    \vspace{1mm}
    \end{minipage} \\
     \hline
     Heuristic & \begin{minipage}{\textwidth}
    \vspace{1mm}
    \raggedleft
    \includegraphics[scale=0.35]{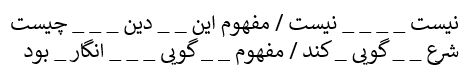}
    \hspace{2.08cm}
    \vspace{1mm}
    \end{minipage} \\
     \hline
     FT & \begin{minipage}{\textwidth}
    \vspace{1mm}
    \raggedleft
    \includegraphics[scale=0.35]{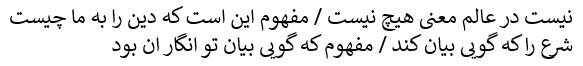}
    \hspace{2.08cm}
    \vspace{1mm}
    \end{minipage}
    \newline
    In this world, what matters the most is belief since the rest is of no importance. It states the right while your words narrate the others. \\
    %  \hline
    %  \hline
    % \multicolumn{2}{c}{\textbf{Sample 14}} \\
    %  \hline
    %  \multirow{ 2}{*}{}Prose & \begin{minipage}{\textwidth}
    % \vspace{1mm}
    % \raggedleft
    % \includegraphics[scale=0.35]{Samples/14_1.JPG}
    % \hspace{2.08cm}
    % \vspace{1mm}
    % \end{minipage} \\
    % \hline
    %  Poem (G) & \begin{minipage}{\textwidth}
    % \vspace{1mm}
    % \raggedleft
    % \includegraphics[scale=0.35]{Samples/14_2.JPG}
    % \hspace{2.08cm}
    % \vspace{1mm}
    % \end{minipage} \\
    %  \hline
    %  IT & \begin{minipage}{\textwidth}
    % \vspace{1mm}
    % \raggedleft
    % \includegraphics[scale=0.35]{Samples/14_3.JPG}
    % \hspace{2.08cm}
    % \vspace{1mm}
    % \end{minipage} \\
    %  \hline
    %  Heuristic & \begin{minipage}{\textwidth}
    % \vspace{1mm}
    % \raggedleft
    % \includegraphics[scale=0.35]{Samples/14_4.JPG}
    % \hspace{2.08cm}
    % \vspace{1mm}
    % \end{minipage} \\
    %  \hline
    %  FT & \begin{minipage}{\textwidth}
    % \vspace{1mm}
    % \raggedleft
    % \includegraphics[scale=0.35]{Samples/14_5.JPG}
    % \hspace{2.08cm}
    % \vspace{1mm}
    % \end{minipage} \\
     \hline
     \hline
    \multicolumn{2}{c}{\textbf{Sample 10}} \\
     \hline
     \multirow{ 2}{*}{}Prose & \begin{minipage}{\textwidth}
    \vspace{1mm}
    \raggedleft
    \includegraphics[scale=0.35]{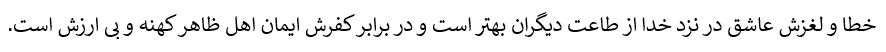}
    \hspace{2.08cm}
    \vspace{1mm}
    \end{minipage} 
    \newline
    Making mistakes as a believer in God is much better than the obedience of the rest, whose faith is ragged and of no value. \\
    \hline
     Poem (G) & \begin{minipage}{\textwidth}
    \vspace{1mm}
    \raggedleft
    \includegraphics[scale=0.35]{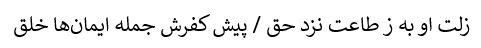}
    \hspace{2.08cm}
    \vspace{1mm}
    \end{minipage} \\
     \hline
     IT & \begin{minipage}{\textwidth}
    \vspace{1mm}
    \raggedleft
    \includegraphics[scale=0.35]{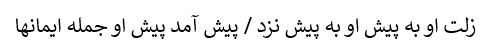}
    \hspace{2.08cm}
    \vspace{1mm}
    \end{minipage} \\
     \hline
     Heuristic & \begin{minipage}{\textwidth}
    \vspace{1mm}
    \raggedleft
    \includegraphics[scale=0.35]{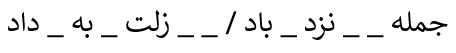}
    \hspace{2.08cm}
    \vspace{1mm}
    \end{minipage} \\
     \hline
     FT & \begin{minipage}{\textwidth}
    \vspace{1mm}
    \raggedleft
    \includegraphics[scale=0.35]{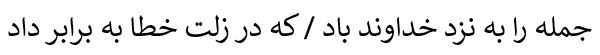}
    \hspace{2.08cm}
    \vspace{1mm}
    \end{minipage}
    \newline
    What I believe in most is the almighty, from whom, if I escape, I will suffer as much penalty as he will have mercy on me if I worship him.\\
    %  \hline
    %  \hline
    % \multicolumn{2}{c}{\textbf{Sample 16}} \\
    %  \hline
    %  \multirow{ 2}{*}{}Prose & \begin{minipage}{\textwidth}
    % \vspace{1mm}
    % \raggedleft
    % \includegraphics[scale=0.35]{Samples/16_1.JPG}
    % \hspace{2.08cm}
    % \vspace{1mm}
    % \end{minipage} \\
    % \hline
    %  Poem (G) & \begin{minipage}{\textwidth}
    % \vspace{1mm}
    % \raggedleft
    % \includegraphics[scale=0.35]{Samples/16_2.JPG}
    % \hspace{2.08cm}
    % \vspace{1mm}
    % \end{minipage} \\
    %  \hline
    %  IT & \begin{minipage}{\textwidth}
    % \vspace{1mm}
    % \raggedleft
    % \includegraphics[scale=0.35]{Samples/16_3.JPG}
    % \hspace{2.08cm}
    % \vspace{1mm}
    % \end{minipage} \\
    %  \hline
    %  Heuristic & \begin{minipage}{\textwidth}
    % \vspace{1mm}
    % \raggedleft
    % \includegraphics[scale=0.35]{Samples/16_4.JPG}
    % \hspace{2.08cm}
    % \vspace{1mm}
    % \end{minipage} \\
    %  \hline
    %  FT & \begin{minipage}{\textwidth}
    % \vspace{1mm}
    % \raggedleft
    % \includegraphics[scale=0.35]{Samples/16_5.JPG}
    % \hspace{2.08cm}
    % \vspace{1mm}
    % \end{minipage} \\
     \hline
\end{longtable}  
\egroup

A number of samples are shown in Table \ref{tab:ourtabel_1} that demonstrate different aspects of our work. Since our poetry dataset includes ambiguous poetical and philosophical texts, they may convey different meanings at the same time. Hence, an exact translation is infeasible. Additionally, there are various figures of speech that further complicate matters. Thus, providing English \textit{equivalents} of these samples was only possible. It is noteworthy that readers should not expect the input and output English equivalents to be precisely the same. There are many ways to illustrate prose's message in poetry, and there is no rule of thumb that can be followed to do so. Even professionals cannot always translate prose directly into poetry. Some words may also seem irrelevant but are figuratively known as emblems that convey certain historical or cultural meanings. Hence, valid judgments of our method's applicability cannot be made from English equivalents which are solely indicated to aid non-native readers to gain insight into Persian prose/poems. Hereafter, we will discuss certain aspects of these examples.

Interestingly, aesthetic values can almost be seen among every FT outcome. For instance, in sample 1, \raisebox{-.2\height}{\includegraphics[height=1.8\fontcharht\font`\B]{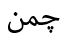}} (Lawn) and \raisebox{-.2\height}{\includegraphics[height=1.8\fontcharht\font`\B]{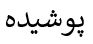}} (covered) are contextually related. Additionally, \raisebox{-.2\height}{\includegraphics[height=1.8\fontcharht\font`\B]{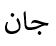}} (spirit), \raisebox{-.2\height}{\includegraphics[height=1.8\fontcharht\font`\B]{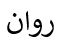}} (mentality), and \raisebox{-.2\height}{\includegraphics[height=1.8\fontcharht\font`\B]{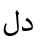}} (soul) are also close in meaning. The two words \raisebox{-.2\height}{\includegraphics[height=1.8\fontcharht\font`\B]{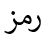}} and \raisebox{-.2\height}{\includegraphics[height=1.8\fontcharht\font`\B]{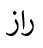}} can also be known as synonyms in Persian literature. Alliteration is another Figure of Speech which consists of words that are similarly written but may or may not differ in a phoneme. In sample 2, the FT consists of the word \raisebox{-.2\height}{\includegraphics[height=1.8\fontcharht\font`\B]{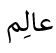}} (wise) and \raisebox{-.2\height}{\includegraphics[height=1.8\fontcharht\font`\B]{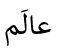}} (world) as two words that are similarly written but differ when are pronounced. The same can also be seen in sample 4 where \raisebox{-.2\height}{\includegraphics[height=1.8\fontcharht\font`\B]{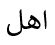}} has been repeated but pronounced similarly, whereas one means docile and the other means inhabitant. Furthermore, a paradox can be seen in the generated poem of sample 5 where \raisebox{-.2\height}{\includegraphics[height=1.8\fontcharht\font`\B]{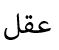}} (intellect) and \raisebox{-.2\height}{\includegraphics[height=1.8\fontcharht\font`\B]{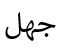}} (foolishness) are antonyms. Phonotactics are known as another figure where a certain character is repeated for several times. In this manner, in sample 6, \raisebox{-.2\height}{\includegraphics[height=1.8\fontcharht\font`\B]{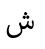}} (pronounced as /sh/) is repeated. These effects demonstrate how different aspects of this work have successfully contributed to poetical values. On the other hand, albeit minor, there are also evidences that depicts how some examples suffer from an increase of prepositions which results in the lack of coherency.

\section{Conclusion}
Persian poetry is well known for its mellifluous poetic style, captivating readers with its unique method of narrating personal, emotional, and historical topics. In computational terms, this mesmerizing phenomenon has solely been investigated as a text generation task. Whereas, in this work, we took the initial steps of modeling Persian poetry as a translation task, where an input Persian prose can be translated to ancient poetry. Our approach involved gathering different datasets and experimenting with encoder-decoder models, masked language modeling, and the beam search algorithm. The two primary challenges in trying to capture centuries of advancements in Persian poetry, are the insufficiency of available parallel data and the highly-technical nature of Persian poems. Hence, we pretrained masked language models using relatively larger monolingual (poem-only) data and incorporated it with a set of heuristics to capture techniques used in generating Persian poetry. By assessing generated poems using official metrics and human assessors, we were able to establish a solid baseline for the Persian prose to poem translation task. 

Future work is expected to integrate more robust methodologies with more technical literary knowledge for prose to poem translation while less supervision and engineering are applied. Most importantly, curating large parallel datasets would certainly enable computational methodologies to gain greater poetic insight, generalize better, and generate higher quality outputs.

%%
%% The next two lines define the bibliography style to be used, and
%% the bibliography file.
\bibliographystyle{ACM-Reference-Format}
\bibliography{manuscript}

\end{document}